\documentclass[10pt,twocolumn,letterpaper]{article}

\usepackage{cvpr}              % To produce the CAMERA-READY version
\usepackage{graphicx}
\usepackage{subfloat}
\usepackage{amsmath}
\usepackage{amssymb}
\usepackage{booktabs}
\usepackage{arydshln}
\usepackage{bbold}
\usepackage{algorithmicx,algorithm}
\usepackage{algorithm}
\usepackage{algpseudocode}
\usepackage{colortbl}
\usepackage{multirow}
\usepackage{array}
\usepackage{caption}
\usepackage{subcaption}
\usepackage{bbding}
\usepackage{pifont}
\usepackage{bbm}
\definecolor{ibmred}{rgb}{0.905,0.435,0.317}
\usepackage[dvipsnames]{xcolor}

\definecolor{cvprblue}{rgb}{0.21,0.49,0.74}
\usepackage[pagebackref,breaklinks,colorlinks,citecolor=cvprblue]{hyperref}

\def\paperID{2516} % *** Enter the Paper ID here
\def\confName{CVPR}
\def\confYear{2024}

\title{Towards Redundancy-Free Sub-networks in Continual Learning}

\author{
Cheng Chen \hspace{-0.2cm}
\and
Jingkuan Song\thanks{Jingkuan Song is the corresponding author.} \hspace{-0.2cm}
\and
LianLi Gao \hspace{-0.2cm}
\and
Heng Tao Shen 
\and \\ 
University of Electronic Science and Technology of China\\ 
{\tt\small \{cczacks,jingkuan.song\}@gmail.com}
}

\begin{document}
\maketitle

%%%%%%%%% ABSTRACT
\begin{abstract}
	Catastrophic Forgetting (CF) is a prominent issue in continual learning.
	Parameter isolation addresses this challenge by masking a sub-network for each task to mitigate interference with old tasks.
	However, these sub-networks are constructed relying on weight magnitude, which does not necessarily correspond to the importance of weights, resulting in maintaining unimportant weights and constructing redundant sub-networks.
	To overcome this limitation, inspired by information bottleneck, which removes redundancy between adjacent network layers, 
	we propose \textbf{\underline{I}nformation \underline{B}ottleneck \underline{M}asked sub-network (IBM)} to eliminate redundancy within sub-networks.
	Specifically, IBM accumulates valuable information into essential weights to construct redundancy-free sub-networks, not only effectively mitigating CF by freezing the sub-networks but also facilitating new tasks training through the transfer of valuable knowledge.
	Additionally, IBM decomposes hidden representations to automate the construction process and make it flexible.
	Extensive experiments demonstrate that IBM consistently outperforms state-of-the-art methods.
	Notably, IBM surpasses the state-of-the-art parameter isolation method with a 70\% reduction in the number of parameters within sub-networks and an 80\% decrease in training time.
	Codes are publicly available at \url{https://github.com/zackschen/IBM-Net}.
	
\end{abstract}

%%%%%%%%% BODY TEXT
\section{Introduction}
\label{intro}
Currently, deep neural networks have demonstrated exceptional performance across various tasks.
Nevertheless, when exposed to new tasks,  they tend to forget previously acquired task-specific information, a phenomenon known as Catastrophic Forgetting \cite{mccloskey1989catastrophic}.
To address this issue,
Continual Learning \cite{DBLP:books/sp/98/Ring98} is proposed to retain previously acquired knowledge while acquiring new information, evolving the networks toward human learning patterns.
In recent years, numerous effective continual learning methods have emerged. 
These methods can be conceptually categorized as follows: regularization-based methods \cite{DBLP:conf/icml/Schwarz0LGTPH18,DBLP:conf/icml/ZenkePG17,DBLP:conf/eccv/AljundiBERT18,DBLP:conf/eccv/LiH16,DBLP:conf/mm/ChenZSG22},
% incorporate additional regularization terms into the loss function to integrate prior knowledge with new information.
memory-based methods \cite{DBLP:conf/iclr/CacciaAATPB22,DBLP:conf/iclr/RiemerCALRTT19,DBLP:conf/nips/BuzzegaBPAC20,DBLP:conf/nips/Lopez-PazR17,DBLP:conf/iclr/ChaudhryRRE19},
%retain previous samples or generate new samples for replay while learning new tasks.
architecture-based methods \cite{DBLP:conf/nips/WortsmanRLKRYF20, DBLP:conf/icml/KangMMYHHY22,DBLP:conf/icml/KonishiKOKK023,DBLP:conf/cvpr/MallyaL18,DBLP:conf/icml/SerraSMK18}.
% create specific modules for each task to mitigate catastrophic forgetting.

\begin{figure}[t]
\centering
\captionsetup[subfloat]{labelsep=none,format=plain,labelformat=empty}
\subfloat[]{
\hspace{-0.1in}
\vspace{-0.2in}
\begin{minipage}[t]{.28\columnwidth}
	\centering
	\includegraphics[width=1\columnwidth]{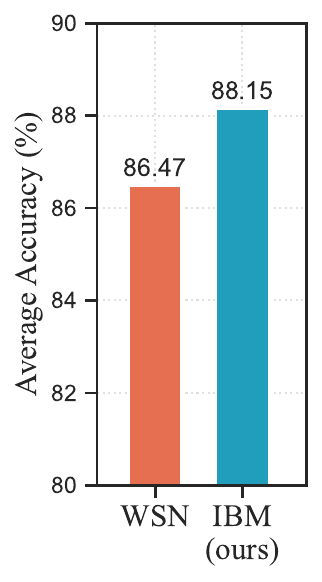}
\end{minipage}
}
\subfloat[]{
\vspace{-0.25in}
\begin{minipage}[t]{0.65\columnwidth}
\centering
\includegraphics[width=1\columnwidth]{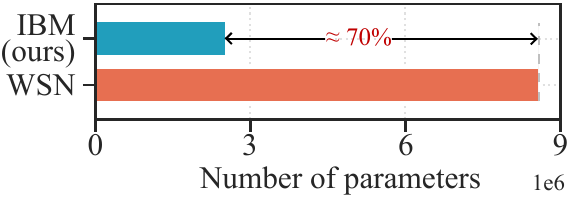} \\
\vspace{0.1in}
\includegraphics[width=1\columnwidth]{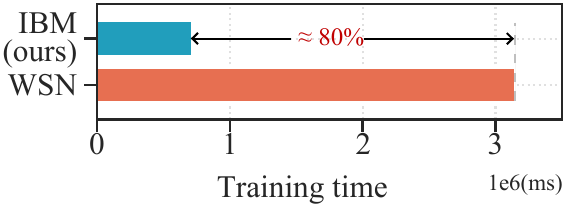} \\
\end{minipage}
}

\caption{The total number of parameters within sub-networks and average training time for each task of WSN \cite{DBLP:conf/icml/KangMMYHHY22} and our IBM on CIFAR-100 with Resnet-18.
Significantly, our method surpasses WSN by 1.68\% with a 70\% reduction in the number of parameters and an 80\% decrease in training time. 
}
\label{fig:intro}
\end{figure}

While existing methods have achieved remarkable success, catastrophic forgetting remains far from being solved.
In the realm of architecture-based methods,
\textit{parameter-isolation} \cite{DBLP:conf/icml/SerraSMK18, DBLP:conf/icml/KangMMYHHY22,DBLP:conf/nips/WortsmanRLKRYF20,DBLP:conf/icml/KonishiKOKK023} methods, which involve compressing or pruning network to construct a specialized sub-network for each task to mitigate interference with old tasks, are arguably the most effective.
%Furthermore, within the realm of architecture-based methods, \textbf{parameter-isolation} \cite{DBLP:conf/icml/SerraSMK18, DBLP:conf/icml/KangMMYHHY22,DBLP:conf/nips/WortsmanRLKRYF20,DBLP:conf/icml/KonishiKOKK023}, which involves compressing or pruning network to create a specialized sub-network for each task, is arguably the most effective approach.
% This approach is based on studies that: deep neural networks tend to be over-parameterized \cite{DBLP:conf/iclr/0022KDSG17, DBLP:conf/nips/HanPTD15}; the removal of redundant or unnecessary weights can yield comparable or even superior performance to the original densely connected network \cite{DBLP:conf/iclr/FrankleC19, DBLP:conf/icml/KangMMYHHY22,DBLP:conf/nips/LinRLZ17}.
%HAT \cite{DBLP:conf/icml/SerraSMK18} and WSN \cite{DBLP:conf/icml/KangMMYHHY22} represent two representative methods within this approach.
%Despite the variations in how these methods construct sub-networks, 
The fundamental criterion of these methods for selecting weights or neurons to construct sub-networks remains roughly the same, i.e., their magnitude.
% HAT learns binary masks for \textit{neurons} specific to each task and blocks the gradient flow through these masks when acquiring new knowledge.
% WSN constructs parameter masks by selecting \textit{weights} with ranking scores in top-$k$ percent.
% These methods construct masks by selecting weights or neurons based on their magnitude.
However, it has been demonstrated by MMC \cite{DBLP:conf/icml/LuoXX22} that the magnitude of weights or neurons does not necessarily correlate with their importance.
Thus, maintaining these unimportant weights or neurons will construct redundancy sub-networks, resulting in an over-consumption of network capacity and a decrease in performance.

Inspired by the success of information bottleneck (\textbf{IB}) theory \cite{DBLP:journals/corr/TishbyZ15,DBLP:conf/icml/DaiZGW18} to penalize an information-theoretic measure of redundancy between adjacent network layers, we conjecture that this ability can be utilized to reduce redundancy within sub-networks, constructing redundancy-free sub-networks.
However, it is not trivial to effectively utilize the IB in the continual learning context because of two critical issues.
Firstly, the original information bottleneck primarily compresses the network with a single task, leaving the question of generalization across multiple tasks open.
Thus, finding an appropriate way to apply IB in the continual learning context is imperative.
Secondly, the IB compression process necessitates a ratio to regulate each layer.
Since lower layers of the network primarily handle visual features while higher layers emphasize semantic information \cite{DBLP:conf/iccv/WangOWL15,DBLP:journals/pami/JingT21}, it is evident that different layers should possess distinct compression ratios.
Previous methods \cite{DBLP:conf/icml/DaiZGW18,DBLP:journals/corr/abs-1712-01312,DBLP:conf/nips/NeklyudovMAV17} set these ratios in a heuristic way manually, which is unsuitable for all tasks or datasets and impractical for deeper networks.

In this work, we bridge the gap by proposing a novel parameter isolation method called \textbf{\underline{I}nformation \underline{B}ottleneck based sub-network \underline{M}ask (IBM)}, a first framework to effectively integrate information bottleneck into continual learning and address these challenges. 
In a nutshell, this method formulates an optimization object by computing inter-layer mutual information, such that valuable information is accumulated into essential parameters while irrelevant information is aggregated into expendable parameters.
Once this optimization process is completed, redundancy-free sub-networks are constructed by selecting these essential parameters.
Further, the key issue in continual learning, \ie catastrophic forgetting, is solved by freezing the essential parameters within sub-networks.
Besides, IBM maintains essential parameters and re-initializes expendable parameters to facilitate knowledge transfer for new tasks learning.
Furthermore, drawing inspiration from recent studies \cite{DBLP:conf/cvpr/GuSS23,DBLP:conf/icml/LuoXX22} that investigate feature distribution and its contribution to final results, we investigate the impact of hidden representation on regulating the compression process.
To this end, a novel feature decomposing module is proposed for IBM to decompose hidden representations to enable automatic and flexible ratio settings.

We conduct extensive experiments on multiple CL benchmarks, which validate our conjecture and demonstrate the effectiveness of IBM.
On both datasets, our method achieves state-of-the-art performance.
Significantly, IBM surpasses WSN (SOTA of parameter-isolation) by 1.68\% with a 70\% reduction in the number of sub-networks parameters and an 80\% decrease in training time, with ResNet-18 on CIFAR-100, as shown in \cref{fig:intro}. 
In summary, the main contributions can be summarized as follows:
\begin{itemize}
	\item 
	%The ability of information bottleneck (IB) to remove redundancy is utilized to reduce redundancy within sub-networks.
	To the best of our knowledge, our work is the first to bring IB into continual learning (CL) context to reduce redundancy within sub-networks.
	
	\item 
	We propose a method IBM to mitigate catastrophic forgetting and facilitate knowledge transfer.
	Additionally, a novel feature detection module is proposed to automate ratio settings for sub-networks construction, allowing for varying ratios across different layers.
	
	\item 
	Extensive experiments confirm the innovative utilization of information bottleneck to reduce redundancy and demonstrate that IBM consistently outperforms SOTA methods while utilizing fewer parameters and training time.
\end{itemize}

\begin{figure*}[t]
	\centering
	\includegraphics[width=0.98\linewidth]{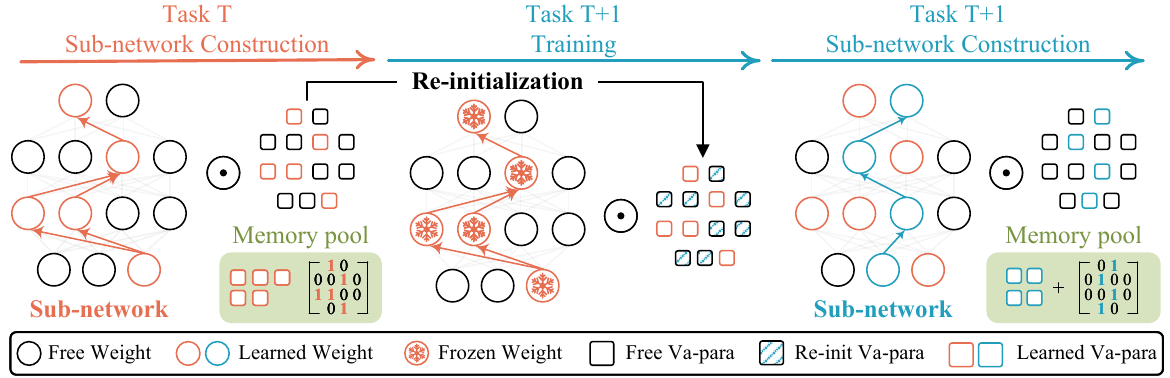}
	\caption{An overview of the proposed IBM.
		After training on Task T, a binary mask is constructed based on choosing a subset of variational parameters.
		The mask and these parameters are copied and saved in a memory pool for inference. 
		Then, before training on T+1, we maintain the chosen parameters and re-initialize the rest parameters to facilitate knowledge transfer.
		Finally, when training on Task T+1, indicated by the masks learned before, the corresponding weights of the network are frozen to solve catastrophic forgetting.
	}
	\label{fig:architecture}
\end{figure*}

\section{Related Work}
Recently, numerous methods have been proposed to mitigate catastrophic forgetting in the continual learning paradigm.
These methods can be broadly categorized into three groups: \textit{regularization-based}, \textit{memory-based} and \textit{architecture-based} methods.

\noindent
\textbf{Regularization-based} methods
\cite{DBLP:conf/icml/Schwarz0LGTPH18,DBLP:conf/icml/ZenkePG17,DBLP:conf/eccv/AljundiBERT18,DBLP:conf/eccv/LiH16,DBLP:conf/mm/ChenZSG22} focus on curing a continual learning network of its catastrophic forgetting by introducing an extra regularization term in the loss function.
\eg, EWC\cite{DBLP:conf/icml/Schwarz0LGTPH18} penalizes the changes of importance parameters when learning new tasks.
%MAS \cite{DBLP:conf/eccv/AljundiBERT18} computes the importance of parameters in an unsupervised and online manner.

\noindent
\textbf{Memory-based} methods \cite{DBLP:conf/iclr/CacciaAATPB22,DBLP:conf/iclr/RiemerCALRTT19,DBLP:conf/nips/BuzzegaBPAC20,DBLP:conf/nips/Lopez-PazR17,DBLP:conf/iclr/ChaudhryRRE19} store previous samples or generate samples for replaying while learning a new task.
Some methods \cite{DBLP:conf/iclr/CacciaAATPB22,DBLP:conf/iclr/RiemerCALRTT19,DBLP:conf/nips/BuzzegaBPAC20,DBLP:conf/nips/BuzzegaBPAC20} use replayed samples from previous tasks to constrain the update of parameters when learning the new task.
During training on a new task of EEC \cite{DBLP:conf/iclr/AyubW21}, reconstructed images from encoded episodes were replayed to avoid catastrophic forgetting.

\noindent
\textbf{Architecture-based} methods \cite{DBLP:conf/nips/WortsmanRLKRYF20, DBLP:conf/icml/KangMMYHHY22,DBLP:conf/icml/KonishiKOKK023,DBLP:conf/cvpr/MallyaL18,DBLP:conf/icml/SerraSMK18} design new architecture modules to each task to prevent any possible forgetting.
PNN \cite{DBLP:journals/corr/RusuRDSKKPH16} adds a network to each task and lateral connections to the network of the previous task while freezing previous task parameters.
MNTDP \cite{DBLP:conf/iclr/VeniatDR21} provides a learning algorithm to search the modules to combine with, where these modules represent atomic skills that can be composed to perform a certain task.

\noindent
\textbf{Parameter-isolation} methods \cite{DBLP:conf/icml/SerraSMK18, DBLP:conf/icml/KangMMYHHY22,DBLP:conf/nips/WortsmanRLKRYF20,DBLP:conf/icml/KonishiKOKK023}, as a branch of architecture-based, try to learn a sub-network for each task by pruning-based or compression-based constraints.
%Piggyback \cite{DBLP:conf/eccv/MallyaDL18} learn a binary mask for each parameter based on an ImageNet pre-trained backbone.
PackNet \cite{DBLP:conf/cvpr/MallyaL18} literately prunes and re-trains the network to construct masks, alleviating catastrophic forgetting by freezing the masked parameter.
However, its performance highly depends on the quality of the pre-trained backbone.
WSN \cite{DBLP:conf/icml/KangMMYHHY22}decouples the learning parameter and the network structure into two separate learnable parameters, weights and weight scores to selecte the weights with top-$k$ percent weight ranking scores.
SPG \cite{DBLP:conf/icml/KonishiKOKK023} computes an importance score for each weight relying on gradients.
The importance score of each parameter is used as a soft-mask to constrain the gradient flow in the backward pass to prevent catastrophic forgetting of previous knowledge.
Although these methods have achieved remarkable performance, they select weights or neurons based on magnitude, resulting in redundant sub-networks. 
Our method is supported by information bottleneck theory to penalize redundancy within sub-networks, achieving redundancy-free sub-networks.

\section{Methodology}
To construct redundancy-free sub-networks, we propose Information Bottleneck Masked sub-network (IBM), a first framework reducing the redundancy of sub-networks in continual learning from an information theory. 
An overview of the IBM is shown in \cref{fig:architecture}.
% In this section, we propose our compression-based method, “\underline{I}nformation \underline{B}ottleneck induced sub-network \underline{M}ask” (IBM). 
% Specifically, we present the problem formulation in continual learning in Section \ref{Statement}. 
% Then, Section \ref{method:IBM} provides the details of the proposed method and Section \ref{method:SVD} illustrates the SVD decomposing method to make the compression automatically and flexibly.

\subsection{Prelimilary}
%\paragraph{Continual Learning Statement}
\label{Statement}
In the setup of supervised continual learning, a series of $T$ tasks are learned sequentially.
The task is denoted by a task descriptor, $\tau \in \{1,2,...,T\}$ and its corresponding dataset $\textit{D}_\tau = \{{(x_{\tau,i},\mathit{y}_{\tau,i} )}_{i=1}^{N_\tau}\}$ which has $N_\tau$ example pairs.
The $x_{\tau}$ ($\in \textit{X}$ ) is the input vector and $\mathit{y}_{\tau}$ ($\in \textit{Y}$ ) is the target vector. 
Let’s consider a $L$ hidden layer neural network, where at each layer network computes the following function to get hidden representation: $h_l = \mathit{f}_l(\mathit{W}_l,h_{l-1})$.
Here, $l \in \{1,2...,L\}$, $\mathit{f}$ is the operation of the network layer and $\mathit{W}$ denotes the layer-wise weight.
At the first layer, $h_0$ = $x_{i}$ represents the raw input data.
This network aims to learn the sequence of tasks by solving the following optimization procedure at each task $\tau$:
\begin{displaymath}
	\Theta^*_\tau = \underset{\Theta}{\textit{minimize}}\frac{1}{N_t} \sum_{i=1}^{N_t}\mathcal{L}_{cl} (\sum_{l=1}^{L} \mathit{f}_l(x_{i};\mathit{W}_l),\mathit{y_{i}} ),
\end{displaymath}
where $\Theta^*_\tau$ denotes the optimal model for task $\tau$ and $\mathcal{L}_{cl}(\cdot,\cdot )$ is a classification objective loss, \eg cross-entropy loss.
It's worth noting that the data $D_{\tau}$ is only accessible when learning for task $\tau$.
In this paper, we mainly focus on the task incremental learning setting, by which task identifier $\tau$ for each data is available during training and testing time.
%Thus the model can focus the decision boundaries within each task.

\subsection{Information Bottleneck Masked sub-network}
\label{method:IBM}

As illustrated in \cite{DBLP:journals/corr/TishbyZ15}, the traditional neural network layers can be interpreted as a Markov chain:
\begin{equation}
	x \to h_1 \to h_2 \to \dots \to h_L \to \hat{y} \to y .
\end{equation}
The hidden layers extract information from previous layers, while the final layer attempts to proximate the true distribution $p(y)$ via output distribution $p(\hat{y}|h_L)$.
Through training the network, sufficient information for accurately predicting $p(y)$ percolates through the network for output.
However, the internal representations will contain superfluous content \cite{DBLP:journals/corr/TishbyZ15, DBLP:conf/icml/DaiZGW18}.
Therefore, the Information bottleneck \cite{DBLP:journals/corr/TishbyZ15} principle penalizes an information-theoretic measure of redundancy between adjacent network layers to identify a mapping $\sum_{l=1}^{L} \mathit{f}_l$ of the input variable that achieves maximum compression while retaining as much information as possible about the output variable.
More specifically, it would like to \textit{minimize} the mutual information $I(h_l;h_{l-1})$, while simultaneously \textit{maximizing} the mutual information $I(h_l;y)$.
Namely, finding an optimal representation $h_l$ is formulated as the minimization of the following Lagrangian:
\begin{equation}
	\label{loss}
	\mathcal{L}_l = \gamma_l I(h_l;h_{l-1}) - I(h_l;y),
\end{equation}
where Lagrangian multiplier $\gamma \ge 0$ serves as a trade-off parameter between the compression of the representation and the amount of preserved relevant information.

Next, we use variational inference to facilitate tractable computation of \cref{loss}.
Specifically, we apply variational distribution $q(h_l)$ and $q(y|h_L)$ to approximate distribution $p(h_l)$ and $p(y|h_L)$, respectively.
Through performing variational inference, we get the upper bound of $\mathcal{L}_l$:
\begin{equation}
	\label{uper_bound}
	\begin{split}
		\mathcal{L}_l \le \hat{\mathcal{L}_l} =  \gamma_l &  \mathbbm{E}_{{h_{l-1}\sim p(h_{l-1}|x)}} [\mathbbm{KL}\left [ p(h_l|h_{l-1})||q(h_l)  \right ]] \\ 
		-& \mathbbm{E}_{h\sim p(h|x)}  [\log q(y|h_L)].
	\end{split}
\end{equation}
Combining all the upper bound $\mathcal{L}_l$ in the network,
$
\mathcal{L}_{all} =\sum_{l}^{} \mathcal{L}_l,
$
minimize $\mathcal{L}_{all}$ is the final variational information bottleneck loss to assimilate information management across all layers.

For dealing with the $p(h_l|h_{l-1})$, $q(h_l)$ and $q(y|h_L)$ in \cref{uper_bound}, a parametric form of these distributions is needed. 
VIBNet \cite{DBLP:conf/icml/DaiZGW18} directly applies the re-parameterization on the hidden representation $h_l$.
However, this operation impedes our effort to integrate IB into the continual learning context through some experimental validation.
Therefore, we opt to use the re-parameterization on the weight of the network:
\begin{equation}
	h_l = f_l(h_{l-1},(\mu_l + \epsilon_l \odot \sigma_l)\odot W_l),
\end{equation}
where $\mu_l$ and $\sigma_l$ are learnable variational parameters (we abbreviate it as \textbf{Va-Para} below) and $\epsilon_l$ is a random parameter sampled from $\mathcal{N}(0,I)$.
The bias in all layers is omitted to avoid clutter.
In this way, we can select the specific weights for each task, facilitating the integration of IBM into the continual learning context.
With the above definitions in mind, the parametric form of $p(h_l|h_{l-1})$ follows that:
\begin{equation}
	\begin{split}
		p(h_l|h_{l-1}) = \mathcal{N}(&h_l;f_l(W_lh_{l-1}\odot \mu_l), \\
		&diag[f_l(W_l^2h_{l-1}^2\odot \sigma _l^2)]. 
	\end{split}
\end{equation}
It is worth noting that Gaussian noise has previously been multiplied with layer-wise activations as a path towards network regularization \cite{DBLP:journals/pami/AchilleS18,DBLP:conf/nips/BlumHP15}.
Proceeding further, we specify the parametric form of $q(h_l)$ following Gaussian distribution as:
\begin{equation}
	q(h_l) = \mathcal{N} (h_l;0,diag[\xi_l]),
\end{equation}
where $\xi_l$ is a variance that can be learned.
These Gaussian assumptions facilitate the creation of an interpretable and closed-form approximation for the KL term in \cref{uper_bound}, enabling us to directly optimize $\xi_l$
Finally, following several algebraic manipulations, we obtain the revised final loss function:
\begin{equation}
	\label{final_loss}
	\mathcal{L}_{all} = {\textstyle \sum_{l=1}^{L}}[\gamma_l \log (1+\frac{\mu^2 }{\sigma^2 } )] - L\mathbbm{E}_{h\sim p(h|x)}[\log q(y|h_L)].
\end{equation}
The trade-off $\gamma$ provides us the flexibility to control the ratio of compression across each layer.

The later term of \cref{final_loss} is a data fit term involving an expectation over latent hidden states, \ie a traditional objective loss.
With respect to the former, it serves as a regularizer based on the KL divergence.
It is noteworthy that the function $\log(1+x)$ is concave and non-decreasing on the domain $[0,\infty]$, representing canonical characteristics of a sparsity-promotion regularizer.
Therefore, rather than favoring a solution with many smaller, partially shrunken versions of the ratios, this type of sparsity archetype:
\begin{equation}
	\alpha_{l,j} = \mu_{l,j}^2\sigma _{l,j}^{-2}, \forall l,j,
\end{equation}
instead prefers pushing some percentages to exactly zero while leaving others mostly unchanged \cite{DBLP:journals/tsp/RaoECPK03,DBLP:conf/icml/DaiZGW18}.
Here, $j$ is the $j$-th element in $\mu_l$ or $\sigma_l$.
Therefore, along with optimizing the object \cref{final_loss}, the useless information will be aggregated into certain expendable weights, as opposed to distributing the information equally across all neurons in a layer \cite{DBLP:conf/icml/DaiZGW18}. 
Once the optimization object converges, a redundancy-free sub-network can be constructed by pruning these expendable weights and selecting the rest weights.
Specifically, we select valuable weights to construct a binary mask matrix for each layer:$M_l = \alpha_{l,j} > 1$ following \cite{DBLP:conf/icml/DaiZGW18}, and save the Va-Para in the memory for inference.
Such that the hidden representation $h_l$ could be recreated:
\begin{equation}
	\begin{split}
		T^{\tau}_l &= (\mu_l + \epsilon_l \odot \sigma_l) \odot  M_l, \\
		h_l &= f(h_{l-1},T^{\tau}_l \odot W_l).
	\end{split}
\end{equation}
It is worth noting that since the activated weights of each layer are far less than the total parameter, the pressure for saving these Va-Para is negligible.
Until now, we have constructed a redundancy-free sub-network for the current task.
The next challenge is how to apply this method within the context of continual learning.

%\subsection{Fitting in Continual learning}
\noindent
\textbf{How to overcome Catastrophic Forgetting.} In the continual learning paradigm, before we train on new tasks, we combine all masks of the previous task:
\begin{equation}
	\begin{split}
		M =  \mathbb{0}, 
		M_{all} =  {\textstyle \sum_{i=1}^{\tau }} M_i || M,
	\end{split}
\end{equation}
where $||$ represents the ‘or’ operation.
The final $M_{all}$ indicates weights that maintain valuable information of previous tasks.
To solve the catastrophic forgetting, we opt to freeze these weights, allowing updates only on the weights that have not been selected.
Specifically, when training on a new task, before performing back-propagation, we modify the gradients $\nabla_{W}\mathcal{L}$ of weights based on the binary masks learned before:
\begin{equation}
	\nabla_{W}\mathcal{L} = \nabla_{W}\mathcal{L} \odot (1-M_{all}),
\end{equation}
effectively freezing the weights selected for the previous tasks.
When performing inference on previous tasks, we can reproduce the same results as before through the frozen weights and the saved Va-Para.

\subsection{Re-initialization of Va-Para} 
However, since the sub-networks for old tasks are frozen, it will limit knowledge transfer for learning new tasks \cite{DBLP:conf/icml/KonishiKOKK023}.
To solve this problem, before training on new tasks, we reuse the Va-Para that is selected by previous masks and re-initialize the rest:
\begin{equation}
	\begin{split}
		\mu_l &= \mu_l \odot M_{all,l} + \mu_{random} \odot (1 - M_{all,l}), \\
		\sigma_l &= \sigma_l \odot M_{all,l} + \sigma_{random} \odot (1 - M_{all,l}).
	\end{split}
\end{equation}
Since the learned Va-Para and the frozen weights accumulate valuable knowledge, the current optimization procedure can select positive knowledge from these parameters and abandon negative one to facilitate current training.
Additionally, re-initialization enables the optimization to escape local optima and discover a more favorable optimization space for future learning.

%With the help of freezing weights and re-initialization, the information bottleneck theory can be applied within the context of continual learning to construct redundancy-free sub-networks, achieving learning with no catastrophic forgetting and knowledge transfer.

\subsection{Feature Decomposing}
\label{method:SVD}
As illustrated in \cref{intro}, since lower layers predominantly handle visual features while higher layers emphasize semantic information,
thus different layers should have different compression ratios.
Thus, we investigate the distribution of hidden representation to set the ratios automatically and flexibly.
In the continual learning community, SVD is a common tool to analyze the hidden representations.
It can be used to factorize hidden representations $H = U \Sigma V^T \in \mathbbm{R}^{m \times n}$ into the product of three matrices, where $U \in \mathbbm{R}^{m \times m}$ and $U \in \mathbbm{R}^{n \times n}$ are orthogonal, and $\Sigma$ contains sorted singular values along its main diagonal \cite{deisenroth2020mathematics}.

In this paper, to investigate the distribution of the hidden representation, we also perform SVD on $h_l = U_l\Sigma_lV_l^T$ followed by its $k$-rank approximation $(h_l)_k$ according to the following criteria for the given threshold $\delta$:
\begin{equation}
	\label{feature_k_rank}
	\left \| (h_l)_k \right \| ^2_F \ge \delta \left \| h_l \right \| ^2_F,
\end{equation}
where $F$ represents Frobenius norm and $\delta$ ($0 <  \delta <  1$) is the threshold hyperparameter.
By applying the $k$-rank approximation, the first $k$ vectors in $U$ represent the significant representation since they contain all the directions with the highest singular values in the representation.
Therefore, the ratio of $k$ over all channels is chosen as the final ratio of compression of layer $l$, \ie $\gamma_i$.
Further, since the feature distribution shifts along the training process, we perform this decomposing process every $E$ epoch to dynamically adjust the ratio, achieving a more precise compression.

\section{Experiment}
%In this section, we first validate the conjecture that using information bottleneck to reduce redundancy within sub-networks.
%Next, we evaluate the effectiveness of our proposed method IBM.
%Last, we perform ablation studies to explore the contribution of each component.

\subsection{Experimental setting}

\paragraph{Datasets}
We conduct experiments on various continual learning benchmarks, including \textbf{CIFAR-100} \cite{krizhevsky2009learning}, \textbf{TinyImageNet}, \textbf{MiniImageNet} \cite{DBLP:conf/nips/VinyalsBLKW16}. 
The CIFAR-100 is constructed by randomly splitting 100 classes of CIFAR-100 into 10 tasks, where each class includes 500 training samples and 100 testing samples.
The TinyImageNet is a variant of the ImageNet dataset \cite{DBLP:conf/cvpr/DengDSLL009} comprising 10 randomized classes out of the 200 classes for each task.
Finally, MiniImageNet is a dataset formed by partitioning the 100 classes of ImageNet into 10 sequential tasks, with each task comprising 10 classes. 
Each class encompasses 500 training samples and 100 testing samples. 

\paragraph{Baselines}
We compare our method with state-of-the-art continual learning methods, including memory-based: \textbf{ER\_ace} \cite{DBLP:conf/iclr/CacciaAATPB22}, \textbf{ER} \cite{DBLP:conf/iclr/RiemerCALRTT19}, \textbf{Der} \cite{DBLP:conf/nips/BuzzegaBPAC20}, \textbf{Der++} \cite{DBLP:conf/nips/BuzzegaBPAC20}, 
\textbf{Gdumb} \cite{DBLP:conf/eccv/PrabhuTD20}, \textbf{Gem} \cite{DBLP:conf/nips/Lopez-PazR17}, \textbf{A-Gem} \cite{DBLP:conf/iclr/ChaudhryRRE19}, and regularization-based:
\textbf{SI} \cite{DBLP:conf/icml/ZenkePG17}, 
\textbf{EWC\_on} \cite{DBLP:conf/icml/Schwarz0LGTPH18}.
Additionally,  three parameter isolation methods are selected, \ie \textbf{Supsup} \cite{DBLP:conf/nips/WortsmanRLKRYF20}, \textbf{WSN} \cite{DBLP:conf/icml/KangMMYHHY22}, \textbf{SPG} \cite{DBLP:conf/icml/KonishiKOKK023}.
Moreover, we consider two simple methods: \textbf{Multi-task} Learning which trains a new network for each task, and \textbf{Finetune} Learning which learns without any mechanism to deal with catastrophic forgetting.

\paragraph{Training setup}
For a fair comparison among CL methods, 
we conduct experiments based on the public library \cite{DBLP:conf/nips/BuzzegaBPAC20}.
Adam optimizer with a fixed learning rate of 0.001 and batch size 256 is applied to all experiments.
We adopt AlexNet \cite{DBLP:conf/nips/KrizhevskySH12} and ResNet-18 \cite{DBLP:conf/cvpr/HeZRS16} as the backbones and train them 300 epochs per task on all datasets.
Besides, we opt to retain 2000 samples for the memory-based methods mentioned above.
For our IBM, the threshold $\delta$ and epoch interval $E$ for feature decomposing is set to 0.97 and 50, respectively.
We conduct each experiment three times with different seeds to obtain averaged metrics.
%Most baselines adopt different hyperparameters for different settings, for which we adopt the hyperparameters in \cite{DBLP:conf/nips/BuzzegaBPAC20} for a fair comparison.
The details can be found in the Appendix.

\paragraph{Evaluation metrics}
We evaluate the performance on the following metrics: Average Accuracy, Backward Transfer and Forward Transfer.
\noindent
(1) \textit{Average Accuracy} (ACC): A average accuracy of all tasks after training on the final task.
$ACC =\frac{1}{T} \sum_{i=1}^TA_{T,i},$
where $A_{T,i}$ is the performance on $\tau=i$ task after training the final task ($\tau=T$).

\noindent
(2) \textit{Backward Transfer} (BWT): This metric measures the catastrophic forgetting after learning on all tasks.
$BWT =\frac{1}{T-1} \sum_{i=1}^{T-1}(A_{T,i}-A_{i,i}),$
where $A_{i, i}$ is the performance on $\tau=i$ task after training on the $\tau=i$ task.

\noindent
(3) \textit{Forward Transfer} (FWT): This metric indicates the positive influence of previous tasks on the current task.
$FWT =\frac{1}{T-1} \sum_{i=1}^{T-1}(A_{i,i}-MT_{i}),$
where $MT_{i}$ is the performance on task $\tau=i$ in Multi-task Learning.

\begin{figure}[t]
\centering
\includegraphics[width=0.9\columnwidth]{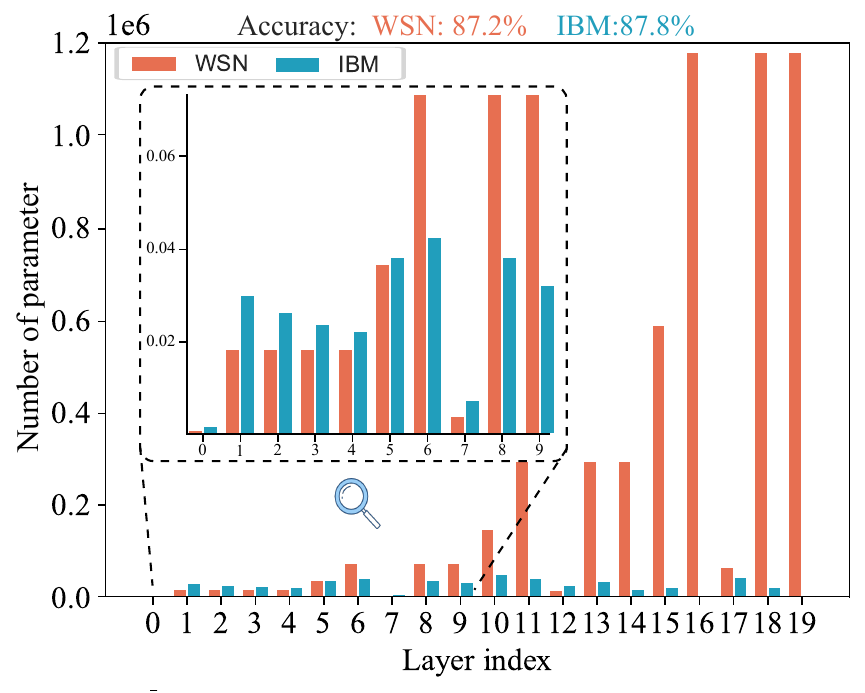} \\
\caption{The visualization of masked weight numbers within each layer of the first sub-network constructed by WSN and IBM on CIFAR-100 with Resnet-18.
The significant reduction of weight numbers substantiates the conjecture of employing the information bottleneck for reducing redundancy within sub-networks.
}
\label{fig:visual_subnetwork}
\end{figure}

\subsection{Conjecture Validation}
Here, we first validate our conjecture that information bottleneck can be utilized for reducing redundancy within sub-networks.
\Cref{fig:visual_subnetwork} shows the masked weights count for each layer of the first sub-network (for the first task), where the results are obtained from WSN and our IBM with ResNet-18 on CIFAR-100.
Here, we choose WSN because it is the parameter-isolation SOTA method now.
In the case of WSN, weight selection involves an equal distribution based on the top-$k$ percent weight ranking scores (set at 50\% in this experiment) across all layers. 
In contrast, our approach allows for dynamic pruning of irrelevant weights and adapts the selection percentage accordingly.
Analysis of the figure reveals that WSN retains a limited number of weights in lower layers but exhibits an abundance of weights in higher layers, due to a significant rise in parameter quantity.
In comparison to WSN, our IBM employs slightly more weights in lower layers but demonstrates a substantial reduction in the number of weights required in higher layers.
This phenomenon can be attributed to the necessity for our IBM to capture ample visual information in the lower layers, allowing it to propagate through the network and provide sufficient semantic information for the higher layers.
Consequently, our IBM achieves a 90\% reduction in the total number of weights and a better performance, surpassing WSN by 0.6\%.
The obtained results provide empirical support for the conjecture that implementing an information bottleneck is effective in reducing redundancy within sub-networks.

\subsection{Comparisons with the State-of-the-arts}
To evaluate the effectiveness of IBM in the continual learning context, we first compare it with state-of-the-art methods with Alexnet on CIFAR-100.
Quantitative results are shown in \cref{tab:alexnet}.
The results show that our IBM achieves the best average accuracy of 82.69\%. 
It is worth noting that, since weights learned by previous tasks have been frozen, our method is proved to be a forget-free model, resulting in 0\% on backward transfer.
In addition, with the help of the re-initialization of Va-Para, IBM exhibits the capability to transfer acquired knowledge to facilitate new task learning, achieving 3.52\% on forward transfer and outperforming previous parameter isolation methods.

\begin{table}[]
	\renewcommand\arraystretch{1.3}
	\renewcommand\tabcolsep{4.0pt}
	\centering
	\resizebox{0.8\columnwidth}{!}{
		\begin{tabular}{lccc}
			\hline
			\multirow{2}{*}{Method}     & \multicolumn{3}{c}{CIFAR-100} \\ \cline{2-4}
			& ACC (\%)  & BWT (\%)  & FWT (\%) \\ \hline
			Multi-task & 79.08$_{\pm0.25}$   & 0        & 0       \\
			Finetune   & 48.97$_{\pm1.59}$   & -38.91$_{\pm1.79}$   & \textbf{4.82}$_{\pm0.46}$    \\ \hline
			ER\_ace \cite{DBLP:conf/iclr/CacciaAATPB22}    & 72.05$_{\pm0.39}$   & -12.37$_{\pm1.59}$   & 4.14$_{\pm0.96}$    \\
			ER \cite{DBLP:conf/iclr/RiemerCALRTT19}         & 71.79$_{\pm0.05}$   & -13.55$_{\pm0.11}$   & 4.82$_{\pm0.35}$    \\
			Der \cite{DBLP:conf/nips/BuzzegaBPAC20}        & 73.27$_{\pm1.08}$   & -10.19$_{\pm0.92}$   & 3.27$_{\pm0.64}$    \\
			Der++ \cite{DBLP:conf/nips/BuzzegaBPAC20}      & 74.12$_{\pm0.72}$   & -10.04$_{\pm1.23}$   & 3.98$_{\pm0.13}$    \\
			SI \cite{DBLP:conf/icml/ZenkePG17}        & 71.67$_{\pm1.14}$   & -10.43$_{\pm0.98}$   & 1.88$_{\pm0.49}$    \\
			EWC\_on \cite{DBLP:conf/icml/Schwarz0LGTPH18}   & 75.33$_{\pm0.54}$   & -0.73$_{\pm0.16}$    & -3.19$_{\pm0.65}$   \\
			AGem \cite{DBLP:conf/iclr/ChaudhryRRE19}      & 65.05$_{\pm1.95}$   & -20.65$_{\pm2.68}$   & 4.47$_{\pm0.33}$    \\
			Gem \cite{DBLP:conf/nips/Lopez-PazR17}       & 74.48$_{\pm0.48}$   & -10.16$_{\pm0.50}$   & 4.45$_{\pm0.33}$        \\
			Gdumb \cite{DBLP:conf/eccv/PrabhuTD20}     & 51.03$_{\pm0.78}$   & -9.71$_{\pm0.83}$    & -19.40$_{\pm0.05}$  \\ \hline
			Supsup \cite{DBLP:conf/nips/WortsmanRLKRYF20}    & 67.30$_{\pm0.66}$   & 0    & -11.83$_{\pm0.67}$       \\
			SPG \cite{DBLP:conf/icml/KonishiKOKK023}       & 69.07$_{\pm0.94}$   & -5.52$_{\pm0.97}$    & -5.17$_{\pm0.44}$       \\
			WSN \cite{DBLP:conf/icml/KangMMYHHY22}       & 82.28$_{\pm0.18}$   & 0        & 3.11$_{\pm0.23}$    \\
			IBM         & \textbf{82.69}$_{\pm0.12}$   & 0        & 3.52$_{\pm0.24}$    \\ \hline
	\end{tabular}}
	\caption{Experimental results of Alexnet backbone on CIFAR-100 dataset.
		The mean and standard deviation of the average accuracy (ACC), backward transfer (BWT) and forward transfer (FWT) across three independent runs are reported. 
		The best results are highlighted in bold.}
	\label{tab:alexnet}
\end{table}

\begin{table*}
	\renewcommand\arraystretch{1.3}
	\renewcommand\tabcolsep{4.0pt}
	\centering
	\resizebox{0.9\linewidth}{!}{
		\begin{tabular}{lccccccccc}
			\hline
			\multirow{2}{*}{Method}     & \multicolumn{3}{c}{CIFAR-100}            & \multicolumn{3}{c}{TinyImageNet}            & \multicolumn{3}{c}{MiniImageNet}            \\ \cline{2-10}
			& ACC (\%)          & BWT (\%)     & FWT (\%) & ACC (\%)          & BWT (\%)         & FWT (\%) & ACC (\%)          & BWT (\%)         & FWT (\%) \\ \hline
			
			Multi-task 
			& 85.16$_{\pm0.54}$ & 0 & 0     
			& 49.98$_{\pm0.35}$ & 0 &  0       
			& 52.69$_{\pm0.25}$ & 0 & 0 \\
			
			Finetune   
			& 34.64$_{\pm2.71}$ & -59.09$_{\pm3.53}$ & 2.66$_{\pm0.84}$         
			& 14.33$_{\pm0.60}$ & -42.11$_{\pm0.45}$ & 2.24$_{\pm0.64}$        
			& 19.63$_{\pm0.51}$ & -38.15$_{\pm0.71}$ & 1.28$_{\pm0.42}$        \\ \hline
			
			ER\_ace \cite{DBLP:conf/iclr/CacciaAATPB22}
			& 77.80$_{\pm0.87}$ & -11.72$_{\pm1.16}$ & 3.18$_{\pm0.72}$         
			& 33.99$_{\pm0.41}$ & -19.21$_{\pm0.58}$ & 1.29$_{\pm0.20}$        
			& 48.01$_{\pm0.64}$ & -8.23$_{\pm0.53}$ & 2.74$_{\pm0.49}$        \\
			
			ER \cite{DBLP:conf/iclr/RiemerCALRTT19}         
			& 77.60$_{\pm0.46}$ & -12.01$_{\pm0.26}$ & 3.25$_{\pm0.14}$        
			& 33.43$_{\pm0.55}$ & -24.30$_{\pm9.24}$ & 0.70$_{\pm0.62}$        
			& 46.91$_{\pm0.66}$ & -8.51$_{\pm0.13}$  & 1.88$_{\pm0.82}$        \\
			
			Der \cite{DBLP:conf/nips/BuzzegaBPAC20}        
			& 83.99$_{\pm0.07}$ & -4.36$_{\pm0.37}$ & 2.75$_{\pm0.30}$        
			& 40.98$_{\pm0.25}$ & -10.97$_{\pm0.73}$ & 0.87$_{\pm1.82}$        
			& 52.45$_{\pm0.37}$ & -3.79$_{\pm0.43}$ & 3.17$_{\pm0.36}$        \\
			
			Der++ \cite{DBLP:conf/nips/BuzzegaBPAC20}   
			& 84.23$_{\pm0.31}$ & -4.58$_{\pm0.59}$ & 3.19$_{\pm0.58}$        
			& 40.70$_{\pm0.53}$ & -11.73$_{\pm0.45}$ & 1.28$_{\pm1.20}$
			& 53.12$_{\pm0.72}$ & -3.81$_{\pm0.37}$ & \textbf{3.86}$_{\pm0.37}$        \\
			
			SI \cite{DBLP:conf/icml/ZenkePG17}
			& 36.87$_{\pm2.47}$ & -54.42$_{\pm2.29}$ & 0.68$_{\pm0.48}$        
			& 14.53$_{\pm0.39}$ & -41.20$_{\pm0.43}$ & 1.63$_{\pm0.64}$        
			& 19.30$_{\pm1.19}$ & -36.97$_{\pm1.71}$ & -0.11$_{\pm1.20}$        \\
			
			EWC\_on \cite{DBLP:conf/icml/Schwarz0LGTPH18}   
			& 74.05$_{\pm0.29}$ & -5.20$_{\pm0.30}$ & -6.44$_{\pm0.96}$        
			& 39.35$_{\pm0.54}$ & -3.96$_{\pm0.22}$ & -7.07$_{\pm0.91}$        
			& 44.14$_{\pm1.32}$ & -7.11$_{\pm0.97}$ & -2.15$_{\pm0.51}$        \\
			
			AGem \cite{DBLP:conf/iclr/ChaudhryRRE19}     
			& 63.16$_{\pm0.57}$ & -28.47$_{\pm0.93}$ & \textbf{3.62}$_{\pm0.87}$        
			& 23.00$_{\pm1.00}$ & -30.73$_{\pm0.61}$ & 0.67$_{\pm0.65}$        
			& 39.26$_{\pm1.38}$ & -16.45$_{\pm1.78}$ & 1.37$_{\pm0.03}$        \\
			
			Gem \cite{DBLP:conf/nips/Lopez-PazR17}       
			& 78.99$_{\pm0.60}$ & -10.31$_{\pm1.21}$ & 3.10$_{\pm0.92}$       
			& 34.79$_{\pm0.29}$ & -19.49$_{\pm0.38}$ & 2.35$_{\pm0.37}$        
			& 47.55$_{\pm0.65}$ & -8.03$_{\pm0.49}$ & 2.09$_{\pm0.47}$        \\
			
			Gdumb \cite{DBLP:conf/eccv/PrabhuTD20}     
			& 42.42$_{\pm0.97}$ & -13.47$_{\pm1.42}$ & -30.62$_{\pm1.35}$        
			& 13.64$_{\pm0.53}$ & -4.46$_{\pm0.58}$ & -32.33$_{\pm0.98}$        
			& 24.21$_{\pm0.63}$ & -7.35$_{\pm0.46}$ & -21.87$_{\pm0.07}$        \\ \hline
			
			SPG \cite{DBLP:conf/icml/KonishiKOKK023}       
			& 48.00$_{\pm1.47}$ & -17.67$_{\pm0.78}$ & -21.20$_{\pm2.09}$       
			& 21.14$_{\pm0.72}$ & -9.15$_{\pm0.28}$ & -20.65$_{\pm1.07}$ 
			& 40.48$_{\pm0.48}$ & -17.78$_{\pm0.30}$ & 3.85$_{\pm0.80}$ \\
			
			Supsup \cite{DBLP:conf/nips/WortsmanRLKRYF20}    
			& 79.94$_{\pm0.41}$ & 0 & -4.97$_{\pm0.26}$        
			& 40.27$_{\pm0.40}$ & 0 & -9.72$_{\pm0.61}$        
			& 50.68$_{\pm0.14}$ & 0 & -2.01$_{\pm0.22}$         \\
			
			WSN \cite{DBLP:conf/icml/KangMMYHHY22}       
			& 86.47$_{\pm0.22}$ & 0 & 1.30$_{\pm0.47}$        
			& 49.36$_{\pm0.34}$ & 0 & -0.62$_{\pm0.30}$        
			& 52.63$_{\pm0.53}$ & 0 & -0.06$_{\pm0.72}$        \\
			
			IBM         
			& \textbf{88.15}$_{\pm0.09}$ & 0 & 2.98$_{\pm0.57}$       
			& \textbf{52.38}$_{\pm0.33}$ & 0 & \textbf{2.40}$_{\pm0.63}$        
			& \textbf{53.90}$_{\pm0.26}$ & 0 & 1.22$_{\pm0.51}$        \\ \hline
		\end{tabular}
	}
	\caption{The quantitative results of each baseline and our IBM based on Resnet-18 backbone.
		We report the mean and standard deviation of the average accuracy (ACC), backward transfer (BWT) and Forward transfer (FWT) across three independent runs with different seeds under the same experimental setup. Our method yields superior results among all datasets.}
	\label{tab:mainResult}
\end{table*}

Then, to further verify the behavior of our IBM on a deeper network, we conduct experiments on the ResNet-18 backbone.
The comparative results on three datasets are presented in \cref{tab:mainResult}.
We have the following observations from this table: \textbf{\underline{Firstly}}, our method consistently achieves the best average accuracy, gaining improvements of 1.68\%, 3.02\%, and 0.78\% compared to SOTA methods on CIFAR-100 (WSN), TinyImageNet (WSN) and MiniImageNet (Der++), respectively.
%of 88.15\%, 52.38\% and 53.90\% in term of ACC on three datasets.
%\textbf{\underline{Secondly}},
%IBM still achieves a forget-free method.
\textbf{\underline{Secondly}},
our method even behaves better than Multi-task which is seen as a strong baseline of continual learning, on all datasets.
The possible reasons are manifolds: 
(1). The compact sub-networks work better than over-parameterized dense neural networks \cite{DBLP:conf/iclr/FrankleC19, DBLP:conf/icml/KangMMYHHY22,DBLP:conf/nips/LinRLZ17}.
After further analysis of the result of each task obtained by IBM and Multi-task, we observe that IBM is always better than Multi-task.
(Details of the results can be found in Appendix.)
(2). The Multi-task lacks the ability to transfer knowledge.
Since it trains a new network for each task, there is no interaction between individual network, which ultimately limits the transferability of knowledge.
In contrast, our method leverages the re-initialization of Va-Para, allowing IBM to selectively retain valuable knowledge and achieve effective knowledge transfer. 
This is evident in the results for FWT, where our approach achieves significant improvements, specifically 2.98\%, 2.40\%, and 1.22\% on three different datasets, respectively.
%In addition, since the SPG uses soft masks to modify the gradient of weights, there is still interference with previous tasks, resulting in suffering from catastrophic forgetting.

\begin{table*}
	\renewcommand\arraystretch{1.3}
	\renewcommand\tabcolsep{4.0pt}
	\centering
	\resizebox{0.9\textwidth}{!}{
		\begin{tabular}{cccccccccc}
			\hline
			\multirow{2}{*}{Variant} & \multirow{2}{*}{FD} & \multicolumn{2}{c}{Re-init} & \multicolumn{2}{c}{CIFAR-100}        & \multicolumn{2}{c}{TinyImageNet}       & \multicolumn{2}{c}{MiniImageNet}      \\ \cline{3-10} 
			&    & Selected        & Unselected        & ACC (\%)           & FWT (\%)          & ACC (\%)            & FWT (\%)           & ACC (\%)           & FWT (\%)           \\ \hline
			\ding{192} & \ding{55}  & \ding{55}           & \ding{55}             & 85.92$_{\pm0.40}$ & 0.75$_{\pm0.27}$ & 48.86$_{\pm0.34}$  & -1.12$_{\pm0.41}$ & 50.32$_{\pm0.17}$ & -2.37$_{\pm0.09}$ \\
			\ding{193} & \textcolor{ibmred}{\checkmark}  & \ding{55}           & \ding{55}             & 86.53$_{\pm0.54}$ & 1.37$_{\pm0.37}$ & 49.49$_{\pm0.33}$  & -0.49$_{\pm0.22}$ & 51.49$_{\pm0.46}$ & -1.20$_{\pm0.48}$ \\
			\ding{194} & \textcolor{ibmred}{\checkmark}  & \ding{55}           & \textcolor{ibmred}{\checkmark}             & 88.15$_{\pm0.09}$ & 2.98$_{\pm0.57}$ & 52.38$_{\pm0.33}$  & 2.40$_{\pm0.63}$  & 53.90$_{\pm0.26}$ & 1.22$_{\pm0.51}$  \\
			\ding{195} & \textcolor{ibmred}{\checkmark}  & \textcolor{ibmred}{\checkmark}           & \textcolor{ibmred}{\checkmark}             & 85.22$_{\pm0.30}$ & 0.05$_{\pm0.27}$ & 47.34$_{\pm0.71}$ & -2.65$_{\pm0.95}$ & 49.93$_{\pm0.92}$ & -2.76$_{\pm1.17}$ \\ \hline
		\end{tabular}
	}
	\caption{The quantitative results for the ablation studies on feature decomposing and re-initialization of IBM, employing Resnet-18 as the backbone.
		It's important to note that since our method does not suffer catastrophic forgetting, we do not record the backward transfer.
		The terms “Selected” and “Unselected” denote the subsets of Va-Para that are chosen by sub-network masks and those that are not, respectively.
	}
	\label{tab:ablation_module}
\end{table*}

\subsection{Ablation studies}
%To deeply investigate IBM, we further study different ablation variants of SVD decomposing and Re-initialization.
%
%To get a clear understanding about the effects of each component,
\noindent
\textbf{Impact of proposed modules.}
Our IBM contains two key components,
including a re-initialization of the Va-Para (Re-init) process for transferring valuable knowledge, as well as a feature decomposing (FD) for adjusting compression ratios.
We conduct component-wise analysis by progressively incorporating them into a baseline model that omits feature decomposing and re-initialization.
Resnet-18 network is chosen as the backbone. 
Since the results in \cref{tab:alexnet} and \cref{tab:mainResult} have confirmed our ability to address catastrophic forgetting, we do not record the backward transfer in these experiments.
The experimental results are shown in \cref{tab:ablation_module}.
From \ding{192} and \ding{193}, when equipped with feature decomposing, we observe a significant increase on average accuracy
(\eg, CIFAR-100: 85.92\% \vs 86.53\%  , TinyImageNet: 48.86\% \vs 49.49\%, 
MiniImageNet: 50.32\% \vs 51.49\%).
This can be construed as the feature decomposing exploring the most suitable ratio for each layer and adjusting these ratios along the training process.
What’s more, these experiment results prove that the feature decomposing not only sets the ratio automatically and flexibly as we designed but also has the ability to promote knowledge transfer, \eg, resulting in a 1.17\% increase on MiniImageNet in terms of FWT.
Additionally, compared \ding{193} and \ding{194}, it can be observed that when training with re-initialization, there is a more substantial improvement on average accuracy (\eg, CIFAR-100: 86.53\% \vs 88.15\%, TinyImageNet: 49.49\% \vs 52.38\%, MiniImageNet: 51.49\% \vs 53.90\%) and forward transfer (\eg, CIFAR-100: 1.37\% \vs 2.98\%, 
TinyImageNet: -0.49\% \vs 2.40\%, MiniImageNet: -1.20\% \vs 1.22\%), demonstrating the effectiveness of re-initialization to extract valuable knowledge while eliminating the negative impact of redundant information.
These results also validate that the re-initialization enables the optimization to escape local optima of previous task and discover a more favorable optimization space for future learning.
Finally, when re-initializing the essential parameters (\ie, \ding{194} and \ding{195}), it yields the worst performances on all datasets.
This occurs because the re-initialization of valuable parameters results in a decline in the acquired knowledge.

\begin{figure}[t]
	\centering
	\begin{minipage}[t]{\columnwidth}
		\centering
		\includegraphics[width=0.9\columnwidth]{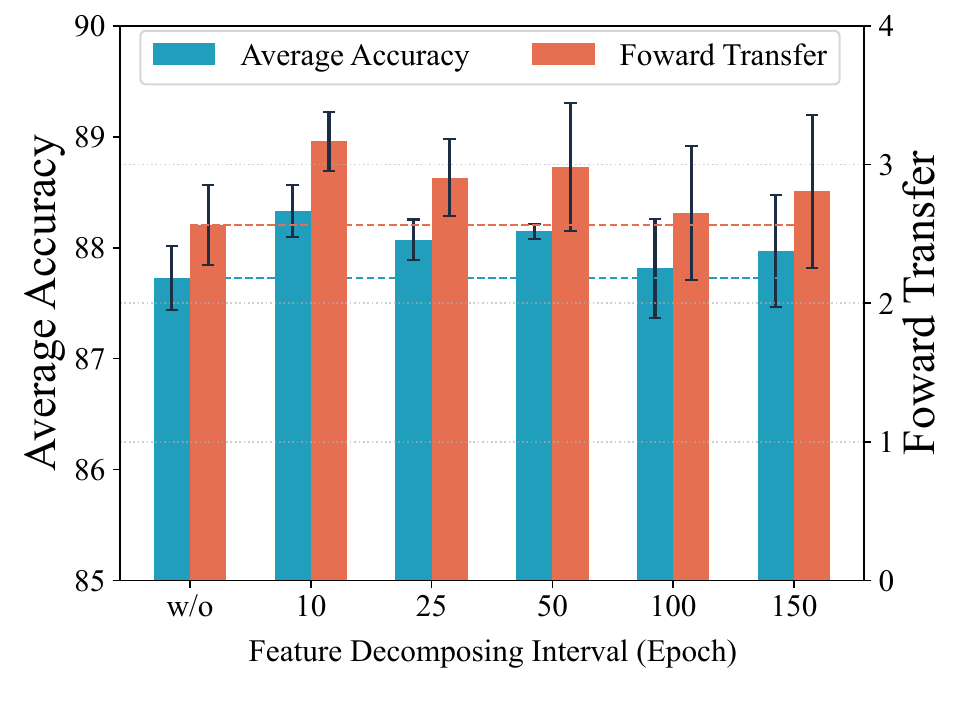} \\
		% \caption{CIFAR-100 Dataset}
		% \includegraphics[width=\columnwidth]{graph/svd_ablation_Tiny.pdf} \\
		% \caption{TinyImageNet Dataset}
		%\includegraphics[width=0.9\columnwidth]{graph/svd_ablation_Mini.pdf} \\
		% \caption{MiniImageNet Dataset}
	\end{minipage}
	\caption{The mean and standard deviation results of ablation studies about feature decomposing interval on CIFAR-100 dataset with ResNet-18 as the backbone.
		We choose 50 epoch interval as the main experiment setting to balance the performance and efficiency.}
	\label{fig:ablation_svd}
\end{figure}

\noindent
\textbf{Impact of feature decomposing frequency.}
To investigate the influence of decomposing interval $E$ in feature decomposing, we conduct experiments on CIFAR-100 with ResNet-18.
In \cref{fig:ablation_svd}, we report the obtained results of the IBM w/o or w/ the feature decomposing module at different training epochs, including 10, 25, 50, 100 and 150. 
(The results of TinyImageNet and MiniImageNet can be found in the Appendix).
In this figure, dashed lines represent the results of \textit{w/o} variant in terms of average accuracy and forward transfer.
As can be observed, all variants with the feature module consistently outperform \textit{w/o} variant, affirming the effectiveness of feature decomposing.
%In addition, what is noteworthy is that the results of all variants with the feature decomposing module are roughly consistent, showing the robustness of feature decomposing under different interval.
In addition, with the decomposing interval increases, the performances decrease.
This trend can be attributed to the fact that, with a smaller interval, the module can more precisely analyze the distribution over time and adjust the compression ratio, leading to enhanced performance.
Nonetheless, a smaller interval demands more training time for analyzing the hidden representation, which limits efficiency.
Hence, to balance performance and efficiency, we opt for a 50-epoch interval as the default setting for the main experiments.

\begin{table}[]
	\renewcommand\arraystretch{1.3}
	\renewcommand\tabcolsep{4.0pt}
	\centering
	\resizebox{0.8\columnwidth}{!}{
		\begin{tabular}{lccc}
			\hline
			\multirow{2}{*}{Method}     & \multicolumn{3}{c}{TinyImageNet} \\ \cline{2-4}
			& ACC (\%)  & BWT (\%)  & FWT (\%) \\ \hline
			Multi-task & 59.14   & 0        & 0       \\
			Finetune   & 16.32   & -49.53   & 4.23    \\ \hline
			ER\_ace \cite{DBLP:conf/iclr/CacciaAATPB22}    & 43.83   & -21.03   & 4.67    \\
			ER \cite{DBLP:conf/iclr/RiemerCALRTT19}        & 44.20    & -20.31   & 4.35    \\
			Der \cite{DBLP:conf/nips/BuzzegaBPAC20}        & 50.86   & -11.78   & 2.91    \\
			Der++ \cite{DBLP:conf/nips/BuzzegaBPAC20}      & 52.88   & -11.62   & 4.78    \\
			SI \cite{DBLP:conf/icml/ZenkePG17}             & 14.35   & -50.85   & 3.52    \\
			EWC\_on \cite{DBLP:conf/icml/Schwarz0LGTPH18}  & 49.45   & -2.62    & -7.20   \\
			AGem \cite{DBLP:conf/iclr/ChaudhryRRE19}       & 25.16   & -38.16   & 2.27    \\
			Gem \cite{DBLP:conf/nips/Lopez-PazR17}         & 48.86   & -16.34   &  \textbf{5.24}       \\
			Gdumb \cite{DBLP:conf/eccv/PrabhuTD20}         & 23.48   & -5.08    & -30.83  \\ \hline
			SPG \cite{DBLP:conf/icml/KonishiKOKK023}       & 16.98 & -5.53 & -36.89       \\
			Supsup \cite{DBLP:conf/nips/WortsmanRLKRYF20}  & 53.83 &  0  &  -5.31  \\
			WSN \cite{DBLP:conf/icml/KangMMYHHY22}         & 61.02   & 0        & 1.88    \\
			IBM                                            & \textbf{63.34}   & 0 & 4.20    \\ \hline
	\end{tabular}}
	\caption{Experimental results of 20-tasks evaluation on TinyImageNet dataset.
		We select Resnet-18 as the backbone.}
	\label{tab:20_task}
\end{table}

\subsection{Capacity for a longer sequence}
A common issue in the parameter isolation is that since the network capacity is limited, “free” parameters tend to saturate as more tasks are introduced, leading to a diminished capacity for learning new tasks.
To assess the capability of our method in overcoming this challenge, we conducted experiments using extended sequences on TinyImageNet, with the quantitative results presented in \cref{tab:20_task} (Additional results for CIFAR-100 and MiniImageNet are included in the Appendix).
As can be seen, our IBM consistently outperforms other methods, without suffering the limited capacity challenge.
This is due to the fact that IBM reduces redundancy in sub-networks, leaving more capacity for future tasks.
This advantage allows IBM to have more capacity to learn fresh knowledge, which leads to better performance.

\section{Conclusion}
This paper proposes a novel parameter-isolation method based on the information bottleneck for pruning the backbone network to construct redundancy-free sub-networks for continual learning.
Moreover, this method freezes sub-networks to mitigate catastrophic forgetting and re-use the essential parameters masked by these sub-networks to transfer valuable knowledge.
In addition, since the different layers of the network process different information, the different layers should hold distinct prune ratios to construct sub-networks.
We propose a feature decomposing module to decompose hidden representations to set the ratio automatically and flexibly.
The results of extensive experiments demonstrate the effectiveness of our method.

%\paragraph{Limitations}

{
	\small
	\bibliographystyle{ieeenat_fullname}
	\bibliography{main}
}

\clearpage
\setcounter{page}{1}
\maketitlesupplementary

\section{Training Detail}
\label{sec:rationale}

For hyper-parameters for other methods, we follow the public library \cite{DBLP:conf/nips/BuzzegaBPAC20}.
The detail hyper-parameters for baselines are shown in \cref{tab:hyper_para}.

\section{Performance on each task of IBM and Multi-task}

The results for each task of IBM and Multi-task on three datasets are presented in \cref{tab:detail_3_datasets}.
Examination of the table reveals that, on the CIFAR-100 dataset, the performance of each task in IBM surpasses that of Multi-task.
This validate the hypothesis that the compact sub-networks work better than over-parameterized dense neural networks \cite{DBLP:conf/iclr/FrankleC19, DBLP:conf/icml/KangMMYHHY22,DBLP:conf/nips/LinRLZ17}.
Furthermore, these results confirm that our IBM possesses the capability to selectively retain valuable knowledge, facilitating effective knowledge transfer.
On more challenging datasets, namely TinyImageNet and MiniImageNet, initially, the performance of IBM is slightly inferior to that of Multi-task.
This can be attributed to the fact that, as task difficulty increases, the sub-networks are not as over-parameterized.
However, when IBM transfers valuable knowledge, this knowledge compensates for the parameter deficiency, leading to improved performance.

\begin{table}[]
	\renewcommand\arraystretch{1.3}
	\renewcommand\tabcolsep{4.0pt}
	\centering
	\resizebox{0.8\columnwidth}{!}{
		\begin{tabular}{ll}
			\hline
			Methods & Hyper-parameter \\ \hline
			ER\_ace & B:2000 \\
			ER      & B:2000 \\
			DER     & B:2000 Mini\_batch\_size:256 Alpha:0.3 \\
			DER++   & B:2000 Mini\_batch\_size:256 Alpha:0.3 Beta:0.5 \\
			SI      & c:10.0 xi:0.1 \\
			EWC\_on & e\_lambda:1000 gamma:10 \\
			AGem    & B:2000 Mini\_batch\_size:256 \\
			Gem     & B:2000 Mini\_batch\_size:256 gamma 0.5 \\
			Gdumb   & B:2000 Mini\_batch\_size:256 Maxlr:5e-4 \\ 
			& Minlr:1e-4 cutmix\_alpha:1 fitting\_epochs:200 \\
			Supsup  & sparsity:8 \\
			WSN     & sparsity:0.5 \\ \hline
	\end{tabular}}
	\caption{List of hyperparameters for the baselines.
		“B” and “Mini\_batch\_size” denote the buffer size and the buffer batch size for rehearsal, respectively.}
	\label{tab:hyper_para}
\end{table}

\section{Decomposing Results on TinyImageNet and MiniImageNet}
\begin{figure}[t]
	\centering
	\begin{minipage}[t]{\columnwidth}
		\centering
		\includegraphics[width=0.9\columnwidth]{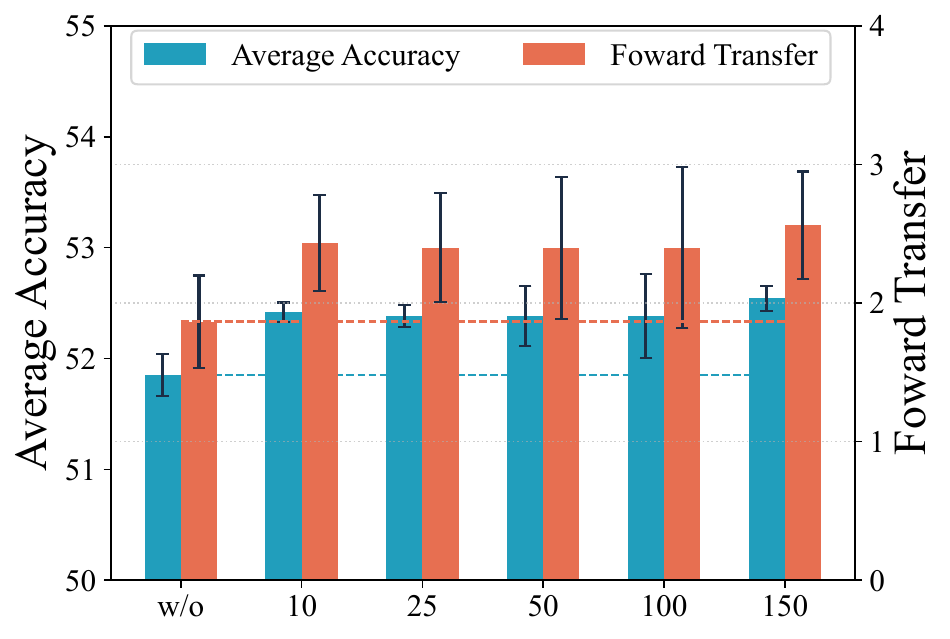} \\
		\includegraphics[width=0.9\columnwidth]{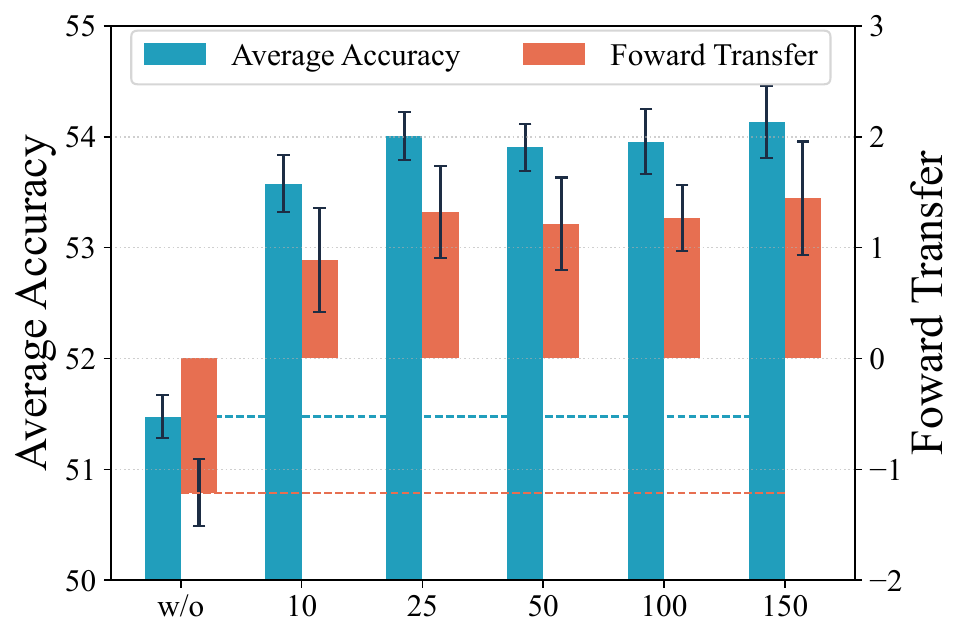} \\
	\end{minipage}
	\caption{The mean and standard deviation results of ablation studies about feature decomposing interval on TinyImageNet and MiniImageNet dataset with ResNet-18 as the backbone.}
	\label{fig:ablation_svd_1}
\end{figure}

We present the performance of feature decomposing variants with different intervals in \cref{fig:ablation_svd}. 
Specifically, \cref{fig:ablation_svd_1} displays the performance of these variants on TinyImageNet and MiniImageNet, utilizing ResNet-18 as the backbone. 
Notably, all variants employing feature decomposing exhibit superior performance compared to the variant without this module, thereby confirming the effectiveness of the feature decomposing module. 
Furthermore, the results obtained with different intervals show a consistent trend on both datasets, TinyImageNet and MiniImageNet, indicating the robustness of our feature decomposing module with respect to varying intervals.

\begin{table}[]
	\renewcommand\arraystretch{1.3}
	\renewcommand\tabcolsep{4.0pt}
	\centering
	\resizebox{\columnwidth}{!}{
		\begin{tabular}{llllllllllll}
			\hline
			Method & Dataset & 0 & 1 & 2 & 3 & 4 & 5 & 6 & 7 & 8 & 9 \\ \hline
			\multirow{3}{*}{IBM} & CIFAR-100 & 87.8 & 88.2 & 86.6 & 88.2 & 88.6 & 86.7 & 89.3 & 86.0 & 90.6 & 90.4 \\
			& TinyImageNet & 45.1 & 42.4 & 51.7 & 53.4 & 56.5 & 60.2 & 49.1 & 57.3 & 56.3 & 49.6 \\
			& MiniImageNet & 53.3 & 56.6 & 50.1 & 52.0 & 52.6 & 58.4 & 58.4 & 51.8 & 51.4 & 55.4 \\ \cdashline{2-12}
			\multirow{3}{*}{Muiti-task} & CIFAR-100 & 86.9 & 83.8 & 85.3 & 87.9 & 83.5 & 83.1 & 84.1 & 80.1 & 84.3 & 88.9 \\
			& TinyImageNet & 46.2 & 44.1 & 49.9 & 51.2 & 53.4 & 57.7 & 47.0 & 53.0 & 54.9 & 46.5 \\
			& MiniImageNet & 54.6 & 53.9 & 49.5 & 49.4 & 53.7 & 56.4 & 55.7 & 47.7 & 47.7 & 56.6 \\ \hline
			%	\cdashline{2-12}
			%	\multirow{3}{*}{WSN} & CIFAR-100 & 87.2 & 84.6 & 85.5 & 88.5 & 84.6 & 85.6 & 87.3 & 84.4 & 86.6 & 91.3 \\
			%	& TinyImageNet & 45.7 & 39.8 & 47.4 & 47.9 & 52.6 & 56.5 & 47.7 & 55.0 & 54.9 & 48.2 \\
			%	& MiniImageNet & 50.2 & 56.4 & 48.3 & 49.8 & 50.7 & 58.0 & 55.5 & 49.9 & 50.0 & 55.8 \\ \hline
		\end{tabular}
	}
	\caption{Quantitative results for each task after training on every task on CIFAR100, TinyImageNet and MiniImageNet.
		These experiments are conducted on Resnet-18 backbone.}
	\label{tab:detail_3_datasets}
\end{table}

\section{Capacity for a longer sequence}
This section provide the results for validating the the capability of our method in overcoming limited capacity, shown in \cref{tab:20_task_cifar} and \cref{tab:20_task_mini}.

\begin{table}[]
	\renewcommand\arraystretch{1.3}
	\centering
	\resizebox{0.8\columnwidth}{!}{
		\begin{tabular}{lccc}
			\hline
			\multirow{2}{*}{Method}     & \multicolumn{3}{c}{CIFAR-100} \\ \cline{2-4}
			& ACC (\%)  & BWT (\%)  & FWT (\%) \\ \hline
			Multi-task & 89.06   & 0        & 0       \\
			Finetune   & 27.47   & -67.82   & 2.84    \\ \hline
			ER\_ace \cite{DBLP:conf/iclr/CacciaAATPB22}    & 81.53   & -10.98   & 2.90    \\
			ER \cite{DBLP:conf/iclr/RiemerCALRTT19}        & 80.92   & -11.07   & 2.38    \\
			Der \cite{DBLP:conf/nips/BuzzegaBPAC20}        & 84.23   & -6.41   & 1.26    \\
			Der++ \cite{DBLP:conf/nips/BuzzegaBPAC20}      & 87.23   & -4.25   & 2.21    \\
			SI \cite{DBLP:conf/icml/ZenkePG17}             & 29.77   & -63.73   & 1.25    \\
			EWC\_on \cite{DBLP:conf/icml/Schwarz0LGTPH18}  & 74.56   & -4.09    & -10.61   \\
			AGem \cite{DBLP:conf/iclr/ChaudhryRRE19}       & 67.04   & -26.13   & 2.80    \\
			Gem \cite{DBLP:conf/nips/Lopez-PazR17}         & 86.26   & -6.35   &  \textbf{3.23}       \\
			Gdumb \cite{DBLP:conf/eccv/PrabhuTD20}         & 55.48   & -10.55    & -23.56  \\ \hline
			SPG \cite{DBLP:conf/icml/KonishiKOKK023}       & 51.75 & -20.79 & -17.56       \\
			Supsup \cite{DBLP:conf/nips/WortsmanRLKRYF20}  & 84.88 &  0  & -4.17  \\
			WSN \cite{DBLP:conf/icml/KangMMYHHY22}         & 90.11   & 0        & 1.05    \\
			IBM                                            & \textbf{90.44}   & 0 & 1.38    \\ \hline
	\end{tabular}}
	\caption{Experimental results of 20-tasks evaluation on CIFAR-100 dataset with ResNet-18.}
	\label{tab:20_task_cifar}
\end{table}

\begin{table}[]
\renewcommand\arraystretch{1.3}
\centering
\resizebox{0.8\columnwidth}{!}{
\begin{tabular}{lccc}
\hline
\multirow{2}{*}{Method}     & \multicolumn{3}{c}{MiniImageNet} \\ \cline{2-4}
& ACC (\%)  & BWT (\%)  & FWT (\%) \\ \hline
Multi-task & 66.5   & 0        & 0       \\
Finetune   & 26.9   & -43.52   & 1.74    \\ \hline
ER\_ace \cite{DBLP:conf/iclr/CacciaAATPB22}    & 62.81   & -7.68   & 3.61   \\
ER \cite{DBLP:conf/iclr/RiemerCALRTT19}        & 61.68    & -8.49   & 3.25    \\
Der \cite{DBLP:conf/nips/BuzzegaBPAC20}        & 63.07   & -5.56   & 1.85    \\
Der++ \cite{DBLP:conf/nips/BuzzegaBPAC20}      & 66.26   & -4.16   & 3.71    \\
SI \cite{DBLP:conf/icml/ZenkePG17}             & 26.45   & -43.73   & 1.49    \\
EWC\_on \cite{DBLP:conf/icml/Schwarz0LGTPH18}  & 58.68   & -3.99    & -4.03   \\
AGem \cite{DBLP:conf/iclr/ChaudhryRRE19}       & 51.36   & -18.14   & 2.09    \\
Gem \cite{DBLP:conf/nips/Lopez-PazR17}         & 63.02   & -7.71   &  \textbf{3.84}       \\
Gdumb \cite{DBLP:conf/eccv/PrabhuTD20}         & 40.97   & -6.66    & -19.2  \\ \hline
SPG \cite{DBLP:conf/icml/KonishiKOKK023}       & 39.88 & -27.53 & -0.35       \\
Supsup \cite{DBLP:conf/nips/WortsmanRLKRYF20}  & 67.34 &  0  & 0.84   \\
WSN \cite{DBLP:conf/icml/KangMMYHHY22}         & 66.9  & 0        & 0.4    \\
IBM                                            & \textbf{68.45}   & 0 & 1.95    \\ \hline
\end{tabular}}
\caption{Experimental results of 20-tasks evaluation on MiniImageNet dataset with ResNet-18.}
\label{tab:20_task_mini}
\end{table}

\end{document}